\documentclass{article}

\usepackage[final]{nips_2017}
\usepackage[utf8]{inputenc}
\usepackage[T1]{fontenc}
\usepackage{hyperref}
\usepackage{url}
\usepackage{booktabs}
\usepackage{amsfonts}
\usepackage{nicefrac}
\usepackage{microtype}
\usepackage{color}
\usepackage{amsmath}
\usepackage{amsthm}
\usepackage{graphicx}
\usepackage{subcaption}
\usepackage{lipsum}
\usepackage{algorithm}
\usepackage{algpseudocode}
\usepackage{xpatch}
\usepackage{fancyvrb}

\makeatletter
\xpatchcmd{\algorithmic}{\ALG@tlm\z@}{\ALG@tlm\z@\leftmargin 0pt}{}{}
\makeatother

\title{A Fixed Version of Quadratic Program in Gradient Episodic Memory}

\author{
  Wei Zhou, Yiying Li\\
  National University of Defense Technology\\
  \texttt{\{zhouwei14,liyiying10\}@nudt.edu.cn} \\
}

\begin{document}

\maketitle

\begin{abstract}
Gradient Episodic Memory (GEM)\citep{lopez2017gradient} is indeed a novel method for continual learning, which solves new problems quickly without forgetting previously acquired knowledge.
However, in the process of studying the paper, we found there were some problems in the proof of the dual problem of Quadratic Program, so here we give our fixed version for this problem.
\end{abstract}

\section{The dual problem of quadratic program}
The main feature of GEM \citep{lopez2017gradient} is an episodic memory $M_t$, which stores a subset of the observed examples from task $t$. In the contribution of GEM, it makes two key observations to solve positive backward transfer efficiently.  First, it is unnecessary to store old predictors $f^{t-1}_\theta$, as long as we guarantee that the loss at previous tasks does not increase after each parameter update $g$.  Second, assuming that the function is locally linear (as it happens around small optimization steps) and that the memory is representative of the examples from past tasks, we can diagnose increases in the loss of previous tasks by computing the angle between their loss gradient vector and the proposed update.
Mathematically, we solve the constraints as:
\begin{equation}
\left\langle g, g_{k} \right\rangle := 
\left
\langle
\frac{\partial \ell(f_\theta(x, t), y)}
{\partial \theta}, 
\frac{\partial \ell(f_\theta, \mathcal{M}_k)}
{\partial \theta}
\right\rangle \ge 0, \mbox{ for all } 
 k < t.
\label{eq:constraints}
\end{equation}
If all the inequality constraints~\eqref{eq:constraints} are satisfied, then
the proposed parameter update $g$ is unlikely to increase the loss at previous
tasks.  On the other hand, if one or more of the inequality
constraints~\eqref{eq:constraints} are violated, then there is at least one
previous task that would experience an increase in loss after the parameter
update.  If violations occur, we propose to project the proposed gradient $g$
to the closest gradient $\tilde{g}$ (in squared $\ell_2$ norm) satisfying all
the constraints~\eqref{eq:constraints}. Therefore, we are interested in: 
\begin{align}
\text{minimize}_{\tilde{g}} \frac{1}{2} \quad& \|g - \tilde{g}\|_2^2\nonumber\\
\text{subject to} \quad& \langle \tilde{g}, g_k \rangle \geq 0 \text{ for all } k < t.\label{eq:gemprimal}
\end{align}

To solve \eqref{eq:gemprimal} efficiently, recall the primal of a Quadratic
Program (QP)\citep{nocedal2006numerical,bot2009duality} with inequality constraints:
\begin{align}
    \text{minimize}_z \quad&\frac{1}{2} z^\top C z + w^\top z\nonumber\\
    \text{subject to} \quad&Az \leq b,\label{eq:primal}
\end{align}
where $C \in \mathbb{R}^{p \times p}$, $w\in \mathbb{R}^p$, $A \in
\mathbb{R}^{(t-1) \times p}$, and $b\in\mathbb{R}^{t-1}$, $p$ is the dimension of gradient vector. 

The Lagrangian dual of a QP is also a QP. We write the Lagrangian function as
\begin{equation}
L(z,v )=\frac{1}{2} z^\top C z +p^\top z+ v^\top (Az-b)
\end{equation}

Defining the (Lagrangian) dual function as $g(v)=\inf _{z}L(z,v )$, we find an infimum of $L$, using $\nabla _{z}L(z,v )=0$ and positive-definiteness of Q:

\begin{equation}
z^{*}=-C^{-1}(A^\top v+w)
\end{equation}

So, the dual problem of~\eqref{eq:primal} 
is:
\begin{align}
    \text{minimize}_{v}   \quad&\frac{1}{2} v^\top AC^{-1}A^\top v + (w^\top C^{-1}A^\top + b^\top) v\nonumber\\
    \text{subject to}   \quad&v \geq 0.\label{eq:dual}
\end{align}

With these notations in hand, we write the primal GEM QP \eqref{eq:gemprimal} as:
\begin{align*}
    \text{minimize}_z   \quad& \frac{1}{2} z^\top z -g^\top z + \frac{1}{2} g^\top g\\
    \text{subject to}   \quad& G z \leq 0,
\end{align*}
where $G = -(g_1, \ldots, g_{t-1})$, and we discard the constant term $g^\top g$.
This is a QP on $p$ variables (the number of parameters of the neural
network), which could be measured in the millions. However, we can pose the
dual of the GEM QP as: 
\begin{align}
    \text{minimize}_{v}   \quad&\frac{1}{2} v^\top G G^\top v - g^\top G^\top v\nonumber\\
    \text{subject to} \quad&v \geq 0,\label{eq:project}
\end{align}
since $z = -G^\top v + g$ and the term $g^\top g$ is constant.  This is a QP on
$t-1 \ll p$ variables, the number of observed tasks so far. Once we solve the
dual problem \eqref{eq:project} for $v^\star$, we can recover the projected
gradient update as $\tilde{g} = -G^\top v^\star  + g$.

\bibliographystyle{plainnat}
\bibliography{example_paper}

\end{document}